\documentclass{article} 
\usepackage{nips12submit_e,times}

\usepackage{graphicx} 
\usepackage{subfigure}

\usepackage{algorithm}
\usepackage{algorithmic}

\usepackage{hyperref}

\usepackage{amsmath}
\usepackage{amsfonts}
\usepackage{amssymb}
\usepackage{amsthm}
\usepackage{graphicx}
\usepackage{epsfig}
\usepackage{bm}
\usepackage{psfrag}

\psfull

\theoremstyle{definition}
\newtheorem{thm}{Theorem}

\newtheorem{defn}{Definition}

\theoremstyle{remark}

\newcommand{\bfgamma}[0]{{ \bm \gamma }}
\newcommand{\bfepsilon}[0]{{ \bm \epsilon }}

\newcommand{\bfphi}[0]{{ \bm \phi }}

\newcommand{\bft}[0]{{ \bm t }}
\newcommand{\bfw}[0]{{ \bm w }}
\newcommand{\bfv}[0]{{ \bm v }}

\newcommand{\bfx}[0]{{ \bm x }}
\newcommand{\bfz}[0]{{ \bm z }}

\newcommand{\bfy}[0]{{ \bm y }}
\newcommand{\bfe}[0]{{ \bm e }}
\newcommand{\bfu}[0]{{ \bm u }}

\newcommand{\myL}[0]{\mathcal{L}}

\title{Dual-Space Analysis of the Sparse Linear Model}

\author{
David Wipf and Yi Wu \\
Visual Computing Group, Microsoft Research Asia \\
\texttt{davidwipf@gmail.com, jxwuyi@gmail.com} \\
}

\nipsfinalcopy

\begin{document}

\maketitle

\begin{abstract}
Sparse linear (or generalized linear) models combine a standard likelihood function with a sparse prior on the unknown coefficients.  These priors can conveniently be expressed as a maximization over zero-mean Gaussians with different variance hyperparameters.  Standard MAP estimation (Type I) involves maximizing over both the hyperparameters and coefficients, while an empirical Bayesian alternative (Type II) first marginalizes the coefficients and then maximizes over the hyperparameters, leading to a tractable posterior approximation.  The underlying cost functions can be related via a dual-space framework from \cite{Wipf11}, which allows both the Type I or Type II objectives to be expressed in either coefficient or hyperparmeter space.  This perspective is useful because some analyses or extensions are more conducive to development in one space or the other.  Herein we consider the estimation of a trade-off parameter balancing sparsity and data fit.  As this parameter is effectively a variance, natural estimators exist by assessing the problem in hyperparameter (variance) space, transitioning natural ideas from Type II to solve what is much less intuitive for Type I.  In contrast, for analyses of update rules and sparsity properties of local and global solutions, as well as extensions to more general likelihood models, we can leverage coefficient-space techniques developed for Type I and apply them to Type II.  For example, this allows us to prove that Type II-inspired techniques can be successful recovering sparse coefficients when unfavorable restricted isometry properties (RIP) lead to failure of popular $\ell_1$ reconstructions. It also facilitates the analysis of Type II when non-Gaussian likelihood models lead to intractable integrations.
\end{abstract}

\section{Introduction}

We begin with the likelihood model
\begin{equation} \label{eq:basic_prob}
\bfy = \Phi \bfx + \bfepsilon,
\end{equation}
\noindent where $\Phi \in \mathbb{R}^{n \times m}$ is a dictionary of unit
$\ell_2$-norm basis vectors, $\bfx \in \mathbb{R}^{m}$ is a vector of unknown coefficients we would like to estimate,
$\bfy \in \mathbb{R}^{n}$ is the observed signal, and $\bfepsilon$ is noise distributed
as $\mathcal{N}(\bfepsilon; 0,\lambda I)$ (later we consider more general likelihood models).  In many practical situations where large numbers of features are present relative to the signal dimension, the problem of estimating $\bfx$ given $\bfy$ becomes ill-posed. A Bayesian framework is intuitively appealing
for formulating these types of problems because prior assumptions must be
incorporated, whether explicitly or implicitly, to regularize the solution space.

Recently, there has been a growing interest in models that employ sparse priors $p(\bfx)$ to
encourage solutions $\bfx$ with mostly small or zero-valued coefficients and a few large or unrestricted values, i.e., we are assuming the generative $\bfx$ is a sparse vector.  Such solutions can be favored by using
\begin{equation} \label{eq:strongly_sg_prior}
p(\bfx) \propto \prod_i \exp\left[-\frac{1}{2} g(x_i) \right] = \prod_i \exp\left[-\frac{1}{2} h\left(x_i^2\right) \right],
\end{equation}
\noindent with $h$ concave and non-decreasing on $[0,\infty)$ \cite{Palmer06,Rao03}.  Virtually all sparse priors of interest can be expressed in this manner, including the popular Laplacian, Jeffreys, Student's $t$, and generalized Gaussian distributions. Roughly speaking, the `more concave' $h$, the more sparse we expect $\bfx$ to be.  For example, with $h(z) = z$, we recover a Gaussian, which is not sparse at all, while $h(z) = \sqrt{z}$ gives a Laplacian distribution, with characteristic heavy tails and a sharp peak at zero.

All sparse priors of the form (\ref{eq:strongly_sg_prior}) can be conveniently framed in terms of a collection of non-negative latent variables or hyperparameters $\bfgamma \triangleq
[\gamma_1,\ldots,\gamma_m]^T$ for purposes of
optimization, approximation, and/or inference. The hyperparameters dictate the structure of the
prior via
\begin{equation} \label{eq:convex_prior}
p(\bfx) = \prod_i p(x_i), \hspace*{0.2cm} p(x_i) = \max_{\gamma_i \geq 0} \mathcal{N}(x_i; 0,\gamma_i) \varphi(\gamma_i),
\end{equation}
\noindent where $\varphi(\gamma_i)$ is some non-negative function that is sometimes treated as a hyperprior, although it will not generally integrate to one. For the purpose of obtaining sparse point estimates of $\bfx$, which will be our primary focus herein, models with latent variable sparse priors are frequently handled in one of two ways.  First, the latent structure afforded by (\ref{eq:convex_prior}) offers a very convenient means of obtaining (possibly local) \textit{maximum a posteriori} (MAP)
estimates of $\bfx$ by iteratively solving 
\begin{equation} \label{eq:type_I_map_problem}
\bfx_{(I)} = \arg\min_\bfx - \log p(\bfy|\bfx)p(\bfx) = \arg\min_{\bfx; \bfgamma \succeq 0} \| \bfy - \Phi \bfx \|_2^2 + \lambda \sum_i \left[\frac{x_i^2}{\gamma_i} + \log \gamma_i + f(\gamma_i)  \right],
\end{equation}
\noindent where $f(\gamma_i) \triangleq -2 \log \varphi(\gamma_i)$ and $\bfx_{(I)}$ is commonly referred to as a \textit{Type I} estimator.  Examples include minimum $\ell_p$-norm approaches \cite{Chartrand08,Kreutz03,Rao03}, Jeffreys prior-based
methods sometimes called FOCUSS \cite{Fevotte06,Figueiredo01,Gorodnitsky97},
algorithms for computing the basis pursuit (BP) or Lasso solution
\cite{Figueiredo01,Rao03,Tibshirani96}, and iterative reweighted $\ell_1$  methods \cite{Candes08}.

Secondly, instead of maximizing over both $\bfx$ and $\bfgamma$ as in (\ref{eq:type_I_map_problem}),
\emph{Type II} methods first integrate out (marginalize) the unknown $\bfx$ and then solve the empirical Bayesian problem \cite{Tipping01}
\begin{eqnarray} \label{eq:type_II_MAP}
\bfgamma_{(II)} & = & \arg\max_\bfgamma p(\bfgamma|\bfy) = \arg\max_\bfgamma \int p(\bfy|\bfx) \prod_i \mathcal{N}(\bfx ; 0,\gamma_i) \varphi(\bfgamma_i) d x_i \nonumber \\
& = & \arg\min_\bfgamma \hspace*{0.2cm} \bfy^T \Sigma_y^{-1}\bfy + \log \left|\Sigma_y \right| + \sum_{i=1}^m f(\gamma_i),
\end{eqnarray}
\noindent where $\Sigma_y \triangleq \lambda I + \Phi \Gamma \Phi^T$ and $\Gamma \triangleq \mbox{diag}[\bfgamma]$.  Once $\bfgamma_{(II)}$ is obtained, the conditional distribution $p(\bfx | \bfy; \bfgamma_{(II)})$ is Gaussian, and a point estimate for $\bfx$ naturally
emerges as the posterior mean
\begin{equation} \label{eq:posterior_mean}
\bfx_{(II)} = \mathrm{E}\left[\bfx|\bfy;\bfgamma_{(II)} \right] = \Gamma_{(II)} \Phi^T
\left(\lambda I + \Phi \Gamma_{(II)} \Phi^T \right)^{-1} \bfy.
\end{equation}
\noindent Pertinent examples include sparse Bayesian learning and the relevance vector machine (RVM) \cite{Tipping01}, automatic relevance
determination (ARD) \cite{Neal96},
methods for learning overcomplete dictionaries \cite{Girolami01}, and large-scale experimental design \cite{Seeger11}.

While initially these two approaches may seem vastly different, both can be directly compared using a dual-space view \cite{Wipf11} of the underlying cost functions.  In brief, this involves expressing both the Type I and Type II objective solely in terms of either $\bfx$ or $\bfgamma$ as reviewed in Section \ref{sec:dual_space_view}.    The dual-space view is advantageous for several reasons, such as establishing connections between algorithms, developing efficient update rules, or handling more general (non-Gaussian) likelihood functions.  In Section \ref{sec:learn_lambda}, we utilize $\bfgamma$-space cost functions to develop a principled method for choosing the trade-off parameter $\lambda$ (which accompanies the Gaussian likelihood model and essentially balances sparsity and data fit) and demonstrate its effectiveness via simulations.  Section \ref{sec:max_sparse_estimation} then derives a new Type II-inspired algorithm in $\bfx$-space that can compute maximally sparse (minimal $\ell_0$ norm) solutions even with highly coherent dictionaries, proving a result for clustered dictionaries that previously has only been shown empirically \cite{Wipf12}.  Finally, Section \ref{sec:gen_likelihood_models} leverages duality to address Type II methods with generalized likelihood functions that previously were rendered untenable because of intractable integrals.  In general, some tasks and analyses are easier to undertake in $\bfgamma$-space (Section \ref{sec:learn_lambda}), while others are more transparent in $\bfx$-space (Sections \ref{sec:max_sparse_estimation} and \ref{sec:gen_likelihood_models}).  Here we consider both with the goal of advancing the proper understanding and full utilization of the sparse linear model.

\section{Dual-Space View of the Sparse Linear Model} \label{sec:dual_space_view}

Type I is based on a natural cost function in $\bfx$-space, $p(\bfx|\bfy)$, while Type II involves an analogous function in $\bfgamma$-space, $p(\bfgamma|\bfy)$.  The dual-space view defines a corresponding $\bfgamma$-space cost function for Type I and a $\bfx$-space cost function for Type II to complete the symmetry.

\textbf{\textit{Type II in $\bfx$-Space}}: Using the relationship
\begin{equation}
\bfy \Sigma_y^{-1} \bfy = \min_x \frac{1}{\lambda} \| \bfy - \Phi \bfx \|_2^2 + \bfx^T \Gamma^{-1} \bfx
\end{equation}
\noindent as in \cite{Wipf11}, it can be shown that the Type II coefficients from (\ref{eq:posterior_mean}) satisfy $\bfx_{(II)} = \arg \min_{\bfx} \myL_{(II)}(\bfx)$, where
\begin{equation} \label{eq:ard_map}
\myL_{(II)}(\bfx) \hspace*{0.2cm} \triangleq \hspace*{0.2cm}  \|\bfy - \Phi \bfx \|_2^2 + \lambda g_{(II)}(\bfx),
\end{equation}
\noindent and
\begin{equation} \label{eq:g_definition}
g_{(II)}(\bfx) \hspace*{0.2cm} \triangleq \hspace*{0.2cm} \min_{\bfgamma \succeq 0} \hspace*{0.2cm} \sum_i \frac{x_i^2}{\gamma_i} + \log |\Sigma_y | + \sum_i f(\gamma_i).
\end{equation}
This reformulation of Type II in $\bfx$-space is revealing for multiple reasons (Sections \ref{sec:max_sparse_estimation} and \ref{sec:gen_likelihood_models} will address additional reasons in detail). For many applications of the sparse linear model, the primary goal is simply a point estimate that exhibits some degree of sparsity, meaning many elements of $\hat{\bfx}$ near zero and a few relatively large coefficients.  This requires a penalty function $g(\bfx)$ that is concave and non-decreasing in $\bfx^2 \triangleq [x_1^2,\ldots,x_m^2]^T$.  In the context of Type I, any prior $p(\bfx)$ expressible via (\ref{eq:strongly_sg_prior}) will satisfy this condition by definition; such priors are said to be \textit{strongly super-Gaussian} and will always have positive kurtosis \cite{Palmer06}.  Regarding Type II, because the associated $\bfx$-space penalty (\ref{eq:g_definition}) is represented as a minimum of upper-bounding hyperplanes with respect to $\bfx^2$ (and the slopes are all non-negative given $\bfgamma \succeq 0$), it must therefore be concave and non-decreasing in $\bfx^2$ \cite{Boyd04}.

For compression, interpretability, or other practical reasons, it is sometimes desirable to have \textit{exactly sparse} point estimates, with many (or most) elements of $\bfx$ equal to exactly zero.  This then necessitates a penalty function $g(\bfx)$ that is concave and non-decreasing in $|\bfx| \triangleq [|x_1|,\ldots,|x_m|]^T$, a much stronger condition.  In the case of Type I, if $\log \gamma + f(\gamma)$ is concave and non-decreasing in $\gamma$, then $g(\bfx) = \sum_i g(x_i)$ satisfies this condition.  The Type II analog, which emerges by further inspection of (\ref{eq:g_definition}) stipulates that if
\begin{equation} \label{eq:concave_requirement}
\log|\Sigma_y| + \sum_i f(\gamma_i) = \log \left|\lambda^{-1}\Phi^T \Phi + \Gamma^{-1} \right| + \log|\Gamma| + \sum_i f(\gamma_i)
\end{equation}
is a concave and non-decreasing function of $\bfgamma$, then $g_{(II)}(\bfx)$ will be a concave, non-decreasing function of $|\bfx|$.  For this purpose it is sufficient, but not necessary, that $f$ be a concave and non-decreasing function.  Note that this is a somewhat stronger criteria than Type I since the first term on the righthand side of (\ref{eq:concave_requirement}) (which is absent from Type I) is actually convex in $\bfgamma$.  Regardless, it is now very transparent how Type II may promote sparsity akin to Type I.

The dual-space view also leads to efficient, convergent algorithms such as iterative reweighted $\ell_1$ minimization and its variants as discussed in \cite{Wipf11}.  However, building on these ideas, we can demonstrate here that it also elucidates the original, widely applied update procedures developed for implementing the relevance vector machine (RVM), a popular Type II method for regression and classification that assumes $f(\gamma) = 0$ \cite{Tipping01}.  In fact these updates, which were inspired by a fixed-point heuristic from \cite{Mackay92}, have been widely used for a number of Bayesian inference tasks without any formal analyses or justification.\footnote{Although a more recent, step-wise variant of the RVM has been shown to be substantially faster \cite{Tipping03}, the original version is still germane since it can easily be extended to handle more general structured sparsity problems.  The step-wise method cannot without introducing additional approximations \cite{Ji09}.}  The dual-space formulation can be leveraged to show that these updates are in fact executing a coordinate-wise, iterative min-max procedure in search of a saddle point.  Specifically we have the following result (all proofs are in the supplementary material):

\vspace*{0.3cm}
\begin{thm} \label{lem:rvm_mackay_updates}
The original RVM update rule from \cite[Equation (16)]{Tipping01} is equivalent to a closed-form, coordinate-wise optimization of
\begin{equation} \label{eq:type_II_regression_expanded2}
\min_{\bfx;\bfgamma \succeq 0} \max_{\bfz \succeq 0} \left[ \|\bfy - \Phi \bfx \|_2^2 + \sum_i \left( \frac{x_i^2}{\gamma_i} + z_i \log \gamma_i \right) - \vartheta(\bfz) \right]
\end{equation}
\noindent over $\bfx$, $\bfgamma$, and $\bfz$, where $\vartheta(\bfz)$ is the convex conjugate function \cite{Boyd04} of $\log \left| \lambda I + \Phi \mbox{diag}[\exp(\bfu)] \Phi^T \right|$ with respect to $\bfu$.
\end{thm}

 \textbf{\textit{Type I in $\bfgamma$-Space}}: Similar methodology and the expansion of $\bfy^T \Sigma_y^{-1} \bfy$ can be used to express the Type I optimization problem in $\bfgamma$-space, which serves several useful purposes. Let $\bfgamma_{(I)} \triangleq \arg\min_{\bfgamma \succeq 0} \myL_{(I)}(\bfgamma)$, with
\begin{equation} \label{eq:type_I_cost}
\myL_{(I)}(\bfgamma) \hspace*{0.2cm} \triangleq \hspace*{0.2cm} \bfy^T \Sigma_y^{-1} \bfy + \log|\Gamma| + \sum_{i=1}^m f(\gamma_i).
\end{equation}
\noindent Then the Type I coefficients obtained from (\ref{eq:type_I_map_problem}) satisfy
\begin{equation} \label{eq:computing_x_I}
\bfx_{(I)} = \Gamma_{(I)} \Phi^T
\left(\lambda I + \Phi \Gamma_{(I)} \Phi^T \right)^{-1} \bfy.
\end{equation}
\noindent  Section \ref{sec:learn_lambda} will use $\bfgamma$-space cost functions to derive well-motivated approaches for learning the trade-off parameter $\lambda$.

\section{Choosing the Trade-off Parameter $\lambda$} \label{sec:learn_lambda}

The trade-off parameter is crucial for obtaining good estimates of $\bfx$.  In general, if $\lambda$ is too large, $\hat{\bfx} \rightarrow 0$; too small and $\hat{\bfx}$ is overfitted to the noise.  In practice, either expensive cross-validation or some heuristic procedure is often required.  However, because $\lambda$ can be interpreted as a variance, it is useful to address its estimation in $\bfgamma$-space, in which existing unknowns (i.e., $\bfgamma$) are also variances.

\textbf{\textit{Learning $\lambda$ with Type I}}: Consider the Type I cost function $\myL_{(I)}(\bfgamma)$.  The data-dependent term can be shown to be a convex, non-increasing function of $\bfgamma$, which encourages each element to be large.  The second term is a penalty factor that regulates the size of $\bfgamma$.  It is here that a convenient regularizer for $\lambda$ can be incorporated.

This can be accomplished as follows. First we expand $\Sigma_y$ via $\Sigma_y = \sum_{j=1}^m \gamma_i \bfphi_{\cdot i} \bfphi_{\cdot i}^T  + \sum_{j=1}^n \lambda \bfe_j \bfe_j^T$, where $\bfphi_{\cdot i}$ denotes the $i$-th column of $\Phi$ and $\bfe_j$ is a column vector of zeros with a `$1$' in the $j$-th location.  Thus we observe that $\lambda$ is embedded in the data-dependent term in the exact same fashion as each $\gamma_i$.  This motivates a penalty on $\lambda$ with similar correspondence, leading to the objective
\begin{eqnarray} \label{eq:type_I_cost_with_lambda}
\myL_{(I)}(\bfgamma,\lambda) & \triangleq & \bfy^T \Sigma_y^{-1} \bfy + \sum_{i=1}^m \left[ \log \gamma_i +
f(\gamma_i) \right] + \sum_{j=1}^n \left[ \log \lambda +
f(\lambda) \right] \nonumber \\
\hspace*{-0.2cm}& = & \bfy^T \Sigma_y^{-1} \bfy + \sum_{i=1}^m \left[ \log \gamma_i +
f(\gamma_i) \right] + n\log \lambda + n f(\lambda).
\end{eqnarray}
\noindent While admittedly simple, this construction is appealing because, regardless of how each $\gamma_i$ is penalized, $\lambda$ is penalized in a proportional manner, so both $\bfgamma$ and $\lambda$ have a properly balanced chance of explaining the observed data.  This is important because the optimal $\lambda$ will be highly dependent on both the true noise level, \textit{and crucially}, the particular sparse prior assumed $p(\bfx)$ (as reflected by $f$).

For analysis or implementational purposes, we may convert $\myL_{(I)}(\bfgamma,\lambda)$ back to $\bfx$-space, with $\lambda$-dependency now removed.  It can then be shown that solving (\ref{eq:type_I_map_problem}), with $\lambda$ fixed to the value that minimizes (\ref{eq:type_I_cost_with_lambda}), is equivalent to solving
\begin{equation} \label{eq:type_I_lambda_x_space}
\min_{\bfx,\bfu} \sum_i g(x_i) + n g\left( \frac{1}{\sqrt{n}} \|\bfu\|_2 \right), \hspace*{0.6cm} \mbox{s.t.} \hspace*{0.3cm} \bfy = \Phi \bfx + \bfu.
\end{equation}
\noindent If $\bfx_*$ and $\bfu_*$ minimize (\ref{eq:type_I_lambda_x_space}), then we can demonstrate using \cite{Palmer06} that the corresponding $\lambda$ estimate, which also minimizes (\ref{eq:type_I_cost_with_lambda}), is given by $\lambda_* = \partial h(z) /\partial z $ evaluated at $z = 1/n \|\bfu_*\|_2^2$.  Note that if we were just performing maximum likelihood estimation of $\lambda$ given $\bfx_*$, the optimal value would reduce to simply $\lambda_* = 1/n \|\bfu_*\|_2^2$, with no influence from the prior on $\bfx$.  This is a fundamental weakness.

Solving (\ref{eq:type_I_lambda_x_space}), or equivalently (\ref{eq:type_I_cost_with_lambda}), can be accomplished using simple iterative reweighted least squares, or if $g$ is concave in $|x_i|$, an iterative reweighted second-order-cone (SOC) minimization.

\textbf{\textit{Learning $\lambda$ with Type II}}: The same procedure can be adopted for Type II yielding the cost function
\vspace*{-0.05cm}
\begin{equation} \label{eq:type_II_cost_with_lambda}
\myL_{(II)}(\bfgamma,\lambda) =  \bfy^T \Sigma_y^{-1} \bfy + \log|\Sigma_y| + \sum_{i}
f(\gamma_i)  + n f(\lambda),
\end{equation}
\noindent where we note that, unlike in the Type I case above, the $\log$-based term is already naturally balanced between $\lambda$ and $\bfgamma$ by virtue of the symmetric embedding in $\Sigma_y$.  It is important to stress that this Type II prescription for learning $\lambda$ is not the same as originally proposed in the literature for Type II models of this genre.  In this context, $\varphi(\gamma_i)$ is interpreted a hyperprior on $\gamma_i$, and an equivalent distribution is assumed on the noise variance $\lambda$.  Importantly, these assumptions leave out the factor of $n$ in (\ref{eq:type_II_cost_with_lambda}), and so an asymmetry is created.

\textbf{\textit{Simulation Examples}}: Empirical tests help to illustrate the efficacy of this procedure.  As in many applications of sparse reconstruction, here we are only concerned with accurately estimating $\bfx$, whose nonzero entries may have physical significance (e.g., source localization \cite{Rao03}, compressive sensing \cite{Candes06}, etc.), as opposed to predicting new values of $\bfy$.  Therefore, automatically learning the value of $\lambda$ is particularly relevant, since cross-validation is often not possible.\footnote{For example, in non-stationary environments, the value of both $\bfx$ and $\lambda$ may be completely different for any new $\bfy$, which then necessitates that we estimate both jointly.}  Simulations are helpful for evaluation purposes since we then have access to the true sparse generating vector.

\begin{figure}[htb]
\psfragscanon
\psfrag{lambda}{$\lambda$ value}
\psfrag{MSE}{\hspace*{0.14cm} MSE}
\psfrag{learned}{ }
\psfrag{  L1penalty}{$\ell_{(0.01)}$}
\psfrag{  L2penalty}{$\ell_{1}$}
\psfrag{sparsity}{$\|\hat{\bfx}\|_0$}
\hspace*{-0.6cm}
\begin{minipage}[b]{.58\linewidth}
  \centering
  \centerline{\epsfig{figure=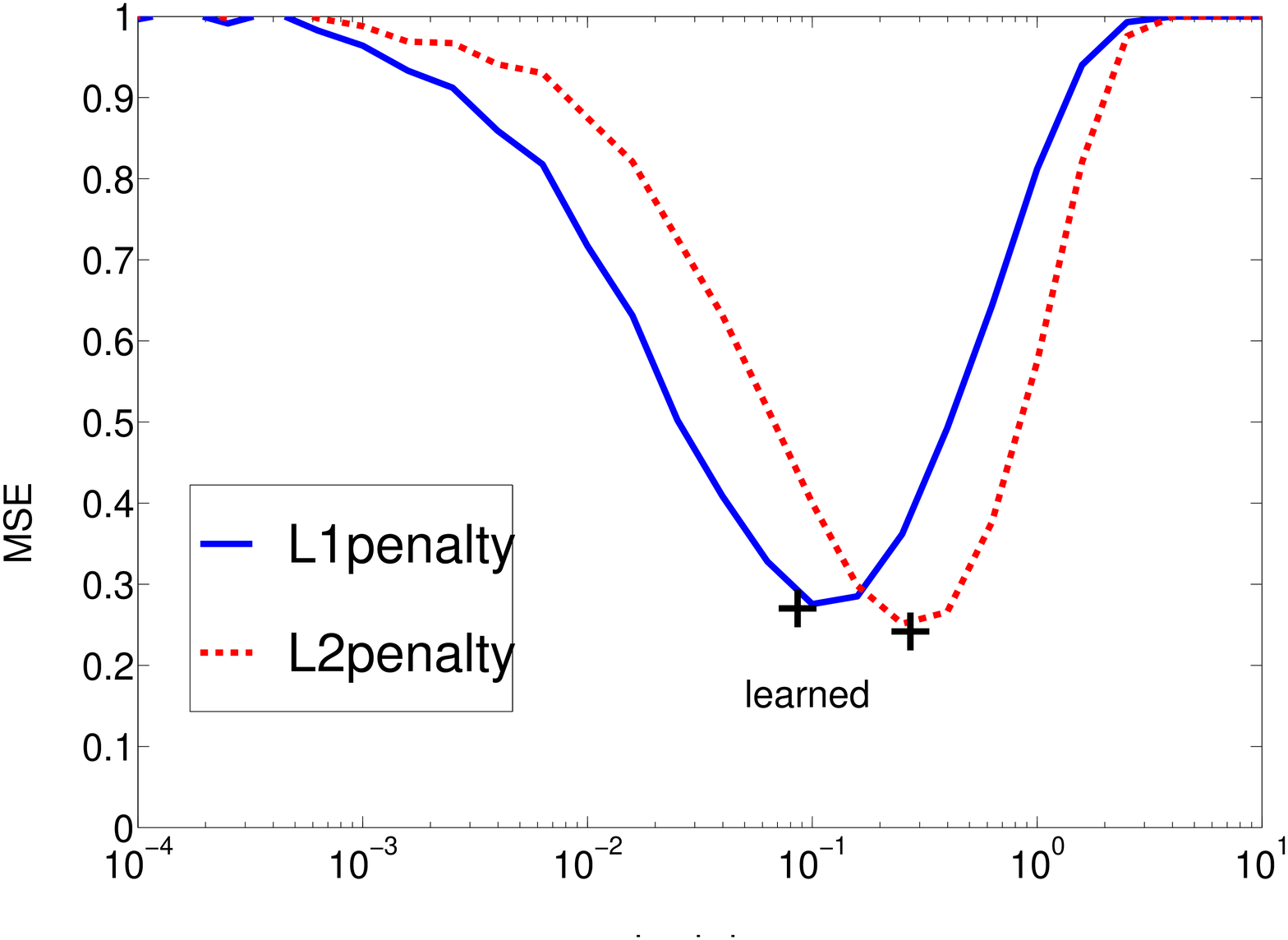,width=7.1cm}}
\end{minipage}
\hfill \hspace*{-0.48cm}
\begin{minipage}[b]{0.53\linewidth}
  \centering
  \centerline{\epsfig{figure=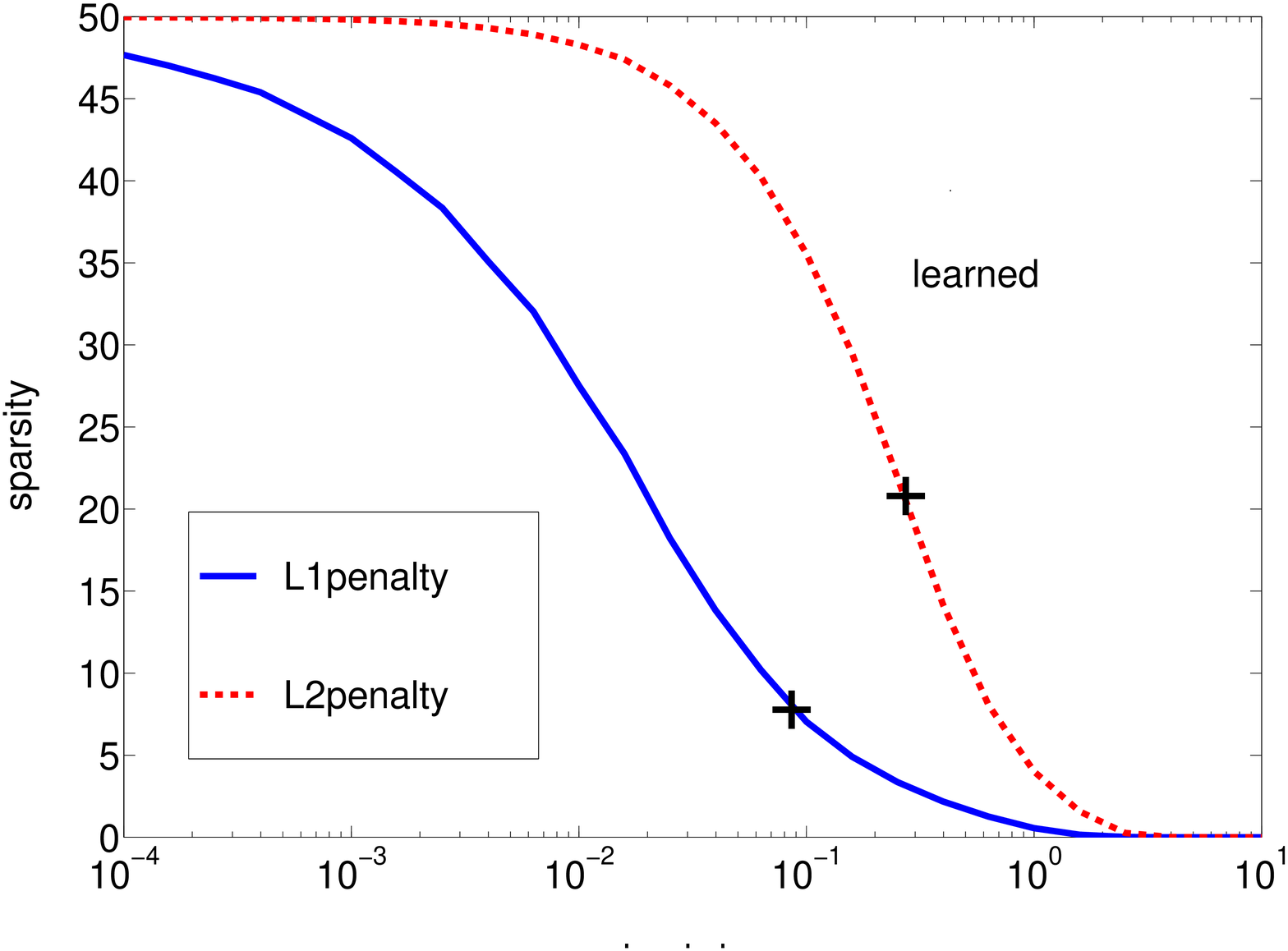,width=7.1cm}}
\end{minipage}
\caption{\emph{Left}: Normalized mean-squared error (MSE) given by $\left< \|\bfx - \hat{\bfx} \|_2^2/\|\bfx \|_2 \right>$ (where the average is across 1000 trials) plotted versus $\lambda$ for two different Type I approaches.  Each black `\textbf{+}' represents the estimated value of $\lambda$ (averaged across trials) and the associated MSE produced with this estimate.  In both cases the estimated value achieves the lowest possible MSE (it can actually be slightly \textit{lower} than the curve because its value is allowed to fluctuate from trial to trial).  \emph{Right}: Solution sparsity $\|\hat{\bfx}\|_0$ versus $\lambda$.  Even though they both lead to similar MSE, the $\ell_{(0.01)}$ penalty produces a much sparser estimate at the optimal $\lambda$ value.} \label{fig:empirical_results}
\end{figure}

Figure \ref{fig:empirical_results} compares the estimation performance obtained by minimizing (\ref{eq:type_I_lambda_x_space}) with two different selections for $g$: $g(\bfx) = \|\bfx\|_p^p = \sum_i |x_i|^p$, with $p = 0.01$ and $p = 1.0$.  Data generation proceeds as follows:  We create a random $100 \times 50$ dictionary $\Phi$, with $\ell_2$-normalized, iid Gaussian columns. $\bfx$ is randomly generated with 10 unit Gaussian nonzero elements.  We then compute $\bfy = \Phi \bfx + \bfepsilon$, where $\bfepsilon$ is iid  Gaussian noise producing an SNR of $0$dB.  To determine what $\lambda$ values lead to optimal performance we solve (\ref{eq:type_I_map_problem}) with the appropriate $g$ over a range of fixed $\lambda$ values ($10^{-4}$ to $10^{1}$) and then compute the error between $\bfx$ and $\hat{\bfx}$.  The minimum of this curve reflects the best performance we can hope to achieve when learning $\lambda$ blindly.  In Figure \ref{fig:empirical_results} (\textit{Top}) we plot these curves  for both Type I methods averaged over 1000 independent trials.

Next we solve (\ref{eq:type_I_lambda_x_space}), which produces an estimate of both $\bfx$ and $\lambda$.  We mark with an `\textbf{+}' the learned $\lambda$ versus the corresponding error of $\hat{\bfx}$.  In both cases the learned $\lambda$'s (averaged across trials) perform just as well as if we knew the optimal value a priori.  Results using other noise levels, problem dimensions $n$ and $m$, sparsity levels $\|\bfx \|_0$, and sparsity penalties $g$ are similar.  See the supplementary material for more examples.

Figure \ref{fig:empirical_results} (\textit{Bottom}) shows the average sparsity of estimates $\hat{\bfx}$, as quantified by the $\ell_0$ norm $\|\hat{\bfx} \|_0$, across $\lambda$ values ($\| \bfx \|_0$ returns a count of the number of nonzero elements in $\bfx$).  The `\textbf{+}' indicates the average sparsity of each $\hat{\bfx}$ for the learned $\lambda$ as before.  In general, the $\ell_{(0.01)}$ penalty produces a much sparser estimate, very near the true value of $\|\bfx\|_0 = 10$ at the optimal $\lambda$.  The $\ell_1$ penalty, which is substantially less concave/sparsity-inducing, still sets some elements to exactly zero, but also substantially shrinks nonzero coefficients in achieving a similar overall reconstruction error.  This highlights the importance of learning a $\lambda$ via a penalty that is properly matched to the prior on $\bfx$: if we instead tried to force a particular sparsity value (in this case 10), then the $\ell_1$ solution would be very suboptimal.  Finally we note that maximum likelihood (ML) estimation of $\lambda$ performs very poorly (not shown), except in the special case where the ML estimate is equivalent to solving (\ref{eq:type_I_cost_with_lambda}) as occurs when $f(\gamma) = 0$ (see \cite{Figueiredo01}).  The proposed method can be viewed as adding a principled hyperprior on $\lambda$, properly matched to $p(\bfx)$, that compensates for this shortcoming of standard ML.

Type II $\lambda$ estimation has been explored elsewhere for the special case where $f(\gamma) = 0$ \cite{Tipping01}, which renders the factor of $n$ in (\ref{eq:type_II_cost_with_lambda}) irrelevant; however, for other selections we have found this factor to improve performance (not shown).  For space considerations we have focused our attention here on Type I, which has frequently been noted for not lending itself well to $\lambda$ estimation (or related parameters) \cite{Figueiredo01,Mattout06}.  In fact, the symmetry afforded by the dual-space perspective reveals that Type I is just as natural a candidate for this task as Type II, and may be preferred in high-dimensional settings where computational resources are at a premium.

\section{Maximally Sparse Estimation} \label{sec:max_sparse_estimation}

With the advent of compressive sensing and other related applications, there has been growing interest in finding \textit{maximally
sparse} signal representations from redundant dictionaries ($m \gg n$)
\cite{Candes08,Donoho03}. The canonical form of this
problem involves solving
\begin{equation} \label{eq:L0_prob}
\bfx_0 \triangleq \arg\min_\bfx \| \bfx \|_0, \hspace*{0.6cm} \mbox{s.t. } \bfy = \Phi \bfx.
\end{equation}
\noindent While (\ref{eq:L0_prob}) is NP-hard, whenever the dictionary $\Phi$ satisfies a
{\em restricted isometry property} (RIP) \cite{Candes06} or a related structural assumption, meaning that
each $\|\bfx_0\|_0$ columns of $\Phi$ are sufficiently close to
orthonormal (i.e., mutually uncorrelated), then replacing $\ell_0$ with $\ell_1$ in (\ref{eq:L0_prob}) leads to a convex problem with an equivalent global solution.  Unfortunately however, in many situations (e.g., feature selection, source localization) these RIP equivalence conditions are grossly violated, implying that the $\ell_1$ solution may deviate substantially from $\bfx_0$.

An alternative is to instead replace (\ref{eq:L0_prob}) with minimization of (\ref{eq:ard_map}) and then take the limit as $\lambda \rightarrow 0$. (Note that the extension to the noisy case with $\lambda > 0$ is straightforward, but analysis is more difficult.) In this regime the optimization problem reduces to
\begin{equation} \label{eq:ard_map_nn}
\bfx_{(II)} = \lim_{\lambda \rightarrow 0} \arg\min_{\bfx} \hspace*{0.1cm} g_{(II)}(\bfx), \hspace*{0.6cm} \mbox{s.t. }
\bfy = \Phi \bfx.
\end{equation}
\noindent  If $\log|\Sigma_y| + \sum_i f(\gamma_i)$ is concave, then (\ref{eq:ard_map_nn}) can be minimized using reweighted $\ell_1$ minimization.  With initial weight vector $\bfw^{(0)} = \mathbf{1}$, the $(k+1)$-th iteration involves computing
\begin{equation} \label{eq:reweighted_L1}
\bfx^{(k+1)}  \leftarrow  \arg\min_{\bfx: \hspace*{0.06cm} \bfy = \Phi \bfx} \sum_i w^{(k)}_i |x_i|, \hspace*{0.8cm} \bfw^{(k+1)}  \leftarrow  \left. \frac{\partial g_{(II)}(\bfx)}{\partial |x_i|}\right|_{\bfx = \bfx^{(k+1)}}.
\end{equation}
\noindent With $f(\gamma) = 0$, iterating (\ref{eq:reweighted_L1}) will provably lead to an estimate of $\bfx_0$ that is as good or better than the $\ell_1$ solution \cite{Wipf12}, in particular when $\Phi$ has highly correlated columns.  Additionally, the assumption $f(\gamma) = 0$ leads to a closed-form expression for the weights $\bfw^{(k+1)}$.  Let
\begin{equation}
\eta_i(\bfx;\alpha,q) \triangleq \left[ \bfphi_{\cdot i}^T \left( \alpha I + \Phi |X^{(k+1)}|^2 \Phi^T  \right)^{-1}  \bfphi_{\cdot i} \right]^{q},
\end{equation}
\noindent where $|X^{(k+1}|$ denotes a diagonal matrix with $i$-th diagonal entry given by $|x_i^{(k+1)}|$.  Then $\bfw^{(k+1)}$ can be computed via $w_i^{(k+1)} = \eta_i(\bfx;0,1/2), \hspace*{0.3cm} \forall i$.  It remains unclear however in what circumstances this type of update can lead to guaranteed improvement nor if the functions $\eta_i(\bfx;0,1/2)$ are even the optimal choice.  We will now demonstrate that for certain selections of $\alpha$ and $q$, we can guarantee that reweighted $\ell_1$ using $\eta_i(\bfx;\alpha, q)$ is guaranteed to recover $\bfx_0$ exactly if $\Phi$ is drawn from what we call a \emph{clustered dictionary model}.
\vspace*{0.3cm}
\begin{defn}
\emph{Clustered Dictionary Model:}  Let $\Phi_{\tiny uncorr}^{(d)}$ denote any dictionary such that $\ell_1$ minimization succeeds in solving (\ref{eq:L0_prob}) for all $\| \bfx_0 \|_0 \leq d$.  Let $\Phi_{\tiny corr}^{(d,\epsilon)}$ denote any dictionary obtained by replacing each column of $\Phi_{\tiny uncorr}^{(d)}$ with a ``cluster'' of $m_i$ basis vectors such that the angle between any two vectors within a cluster is less than some $\epsilon > 0$.  We also define the cluster support $\Omega_0 \subset \{1,2,\ldots,m \}$ as the set of cluster indices whereby $\bfx_0$ has at least one nonzero element.  Finally, we assume that the resulting $\Phi_{\tiny corr}^{(d,\epsilon)}$ is such that every $n \times n$ submatrix is full rank.
\end{defn}
\vspace*{0.2cm}
\begin{thm} \label{thm:irL1_recovery}
For any sparse vector $\bfx_0$ and any dictionary $\Phi_{\tiny corr}^{(d,\epsilon)}$ obtained from the clustered dictionary model with $\epsilon$ sufficiently small, reweighted $\ell_1$ minimization using weights $\eta_i(\bfx;\lambda,q)$ with some $q \geq 1$ and $\alpha$ sufficiently small will recover $\bfx_0$ exactly provided that $|\Omega_0| \leq d$, $\hspace*{0.2cm}$ $\sum_{i\in \Omega_0} m_i \leq n$, and within each cluster $k \in \Omega_0$ the coefficients do not sum to zero.
\end{thm}
\vspace*{0.1cm}

Theorem \ref{thm:irL1_recovery} implies that even though $\ell_1$ may fail to find the maximally sparse $\bfx_0$ because of severe RIP violations (high correlations between groups of dictionary columns as dictated by $\epsilon$ lead directly to a poor RIP), a Type II-inspired method can still be successful.  Moreover, because whenever $\ell_1$ does succeed, Type II will always succeed as well (assuming a reweighted $\ell_1$ implementation), the converse (RIP violation leading to Type II failure but not $\ell_1$ failure) can never happen.  Recent work from \cite{Wipf12} has argued that Type II may be useful for addressing the sparse recovery problem with correlated dictionaries, and empirical evidence is provided showing vastly superior performance on clustered dictionaries.  However, we stress that no results proving global convergence to the correct, maximally sparse solution have been shown before in the case of structured dictionaries (except in special cases with strong, unverifiable constraints on coefficient magnitudes \cite{Wipf12}).  Moreover, the proposed weighting strategy $\eta_i(\bfx;\lambda,q)$ accomplishes this without any particular tuning to the clustered dictionary model under consideration and thus likely holds in many other cases as well.

\section{Generalized Likelihood functions} \label{sec:gen_likelihood_models}

Type I methods naturally accommodate alternative likelihood functions.  We simply must replace the quadratic data fit term from (\ref{eq:type_I_map_problem}) with some preferred function and then coordinate-wise optimization may proceed provided we have an efficient means of computing a weighted $\ell_2$-norm penalized solution.  In contrast, generalizing Type II is substantially more complicated because it is no longer possible to compute the marginalization (\ref{eq:type_II_MAP}) or the posterior distribution $p(\bfx|\bfy;\bfgamma_{(II)})$.  Therefore, to obtain a tractable estimate $\bfx_{(II)}$ additional heuristics are required.  For example, the RVM classifier from \cite{Tipping01} employs a Laplace approximation for this purpose; however, it is not clear what cost function is being minimized nor rigorous properties of the estimated solutions.


Fortunately, the dual $\bfx$-space view provides a natural mechanism for generalizing the basic Type II methodology to address alternative likelihood functions in a more principled manner.  In the case of classification problems, we might want to replace the Gaussian likelihood $p(\bfy|\bfx)$ implied by (\ref{eq:basic_prob}) with a multivariate Bernoulli distribution $p(\bfy|\bfx) \propto \log[-\psi(\bfy,\bfx)]$ where $\psi(\bfy,\bfx)$ is the function
\begin{equation}
\psi\left(\bfy,\bfx \right) \triangleq \sum_j \left( y_j \log \left[ \sigma_j(\bfx) \right] + (1 - y_j) \log \left[ 1 - \sigma_j(\bfx) \right] \right).
 \end{equation}
\noindent Here $y_j \in \{0,1\}$ and $\sigma_j(\bfx) \triangleq 1/[1+\exp(\bfphi_{j \cdot}^T \bfx)]$, with $\bfphi_{j \cdot}$ denoting the $j$-th row of $\Phi$.  This function may be naturally substituted into the $\bfx$-space Type II cost function (\ref{eq:ard_map}) giving us the candidate penalized logistic regression function
\begin{equation} \label{eq:type_II_classifier}
\min_{\bfx}  \hspace*{0.1cm} \psi\left(\bfy,\bfx \right) + \lambda g_{(II)}(\bfx).
\end{equation}
\noindent Importantly, recasting Type II classification using $\bfx$-space in this way, with its attendant well-specified cost function, facilitates more concrete analyses (see below) regarding properties of global and local minima that were previously rendered inaccessible because of intractable integrals and compensatory approximations.  Moreover, we retain a tight connection with the original Type II marginalization process as follows.

Consider the strict upper bound on the function $\psi(\bfy,\bfx)$ (obtained by a Taylor series approximation and a Hessian bound) given by
\begin{equation}
 \psi(\bfy,\bfx) \leq  \pi(\bfy,\bfx,\bfv) \triangleq \psi(\bfy,\bfv) +  \left(\bfv - \bfx \right)^T \Phi^T\bft + 1/8 \left(\bfv - \bfx \right)^T  \Phi^T \Phi \left(\bfv - \bfx \right),
\end{equation}
\noindent where $\bft = [t_1,\ldots,t_n]^T$ with $t_j \triangleq y_j - \sigma_j(\bfv)$. This bound holds for all $\bfv$ with equality when $\bfv = \bfx$.  Using this result we obtain the lower bound on the marginal likelihood given by $\int \log[ - \psi(\bfy,\bfx)] p(\bfx) d\bfx \geq \int \log[- \pi(\bfy,\bfx,\bfv)] p(\bfx) d\bfx$.  The dual-space framework can then be used to derive the following result:
\vspace*{0.3cm}
\begin{thm} \label{thm:rvm_class_bound}
Minimization of (\ref{eq:type_II_classifier}) with $\lambda = 4$ is equivalent to solving
\begin{equation} \label{eq:rvm_class_bound}
\max_{\bfv; \bfgamma \succeq 0} \int \exp\left[-\pi(\bfy,\bfx,\bfv)\right] \prod_i \mathcal{N}(\bfx ; 0,\gamma_i) \varphi(\bfgamma_i) d x_i
\end{equation}
\noindent and then computing $\bfx_{(II)}$ by plugging the resulting $\bfgamma$ into (\ref{eq:posterior_mean}).  
\end{thm}
\vspace*{0.2cm}
\noindent Thus we may conclude that (\ref{eq:type_II_classifier}) provides a principled approximation to (\ref{eq:type_II_MAP}) when a Bernoulli likelihood function is used for classification purposes.  In empirical tests on benchmark data sets (see supplementary material) using $f(\gamma) = 0$, it performs nearly identically to the original RVM (which also implicitly assumes $f(\gamma) = 0$), but nonetheless provides a more solid theoretical justification for Type II classifiers because of the underlying similarities and identical generative model.  But while the RVM and its attendant approximations are difficult to analyze, (\ref{eq:type_II_classifier}) is relatively transparent.    Additionally, for other sparse priors, or equivalently other selections for $f$, we can still perform optimization and analyze cost functions without any conjugacy requirements on the implicit $p(\bfx)$.

\vspace*{0.3cm}
\begin{thm} \label{thm:generalized_likelihood}
If $\log|\Sigma_y| + \sum_i f(\gamma_i)$ is a concave, non-decreasing function of $\bfgamma$ (as will be the case if $f$ is concave and non-decreasing), then every local optimum of (\ref{eq:rvm_class_bound}) is achieved at a solution with at most $n$ nonzero elements in $\bfgamma$ and therefore $\bfx_{(II)}$.  In contrast, if $-\log p(\bfx)$ is convex, then (\ref{eq:rvm_class_bound}) can be globally solved via a convex program. 
\end{thm}
\vspace*{0.2cm}

Despite the practical success of the RVM and related Bayesian techniques, and empirical evidence of sparse solutions, there is currently no proof that the standard variants of these classification methods will always produce exactly sparse estimates.  Thus Theorem \ref{thm:generalized_likelihood} provides some analytical validation of these types of classifiers.

Finally, if we take (\ref{eq:type_II_classifier}) as our starting point, we may naturally consider modifications tailored to specific sparse classification tasks (that may or may not retain an explicit connection with the original Type II probabilistic model).  For example, suppose we would like to obtain a maximally sparse classifier, where regularization is provided by a $\|\bfx\|_0$ penalty.  Direct optimization is combinatorial because of what we call the \emph{global zero attraction property}:  Whenever any individual coefficient $x_i$ goes to zero, we are necessarily at a local minimum with respect to this coefficient because of the infinite slope (discontinuity) of the $\ell_0$ norm at zero.  However, (\ref{eq:type_II_classifier}) can be modified to approximate the $\ell_0$ without this property as follows.
\vspace*{0.3cm}
\begin{thm}
Consider the Type II-inspired minimization problem
\begin{equation} \label{eq:approx_L0_classifier}
\hat{\bfx},\hat{\bfgamma} = \arg\min_{\bfx;\bfgamma \succeq 0} \psi\left(\bfy,\bfx \right) + \alpha_1 \sum_i \frac{x_i^2}{\gamma_i} + \log\left|\alpha_2 I + \Phi \Gamma \Phi^T \right|
\end{equation}
\noindent which is equivalent to (\ref{eq:type_II_classifier}) with $f(\gamma) = 0$ when $\alpha_1 = \alpha_2 = \lambda$.  For some $\alpha_1$ and $\alpha_2$ sufficiently small (but not necessarily equal), the support\footnote{\emph{Support} refers to the index set of the nonzero elements.} of $\hat{\bfx}$ will match the support of $\arg\min_{\bfx} \psi\left(\bfy,\bfx \right) + \lambda \|\bfx\|_0$.  Moreover, (\ref{eq:approx_L0_classifier}) does \emph{not} satisfy the global zero attraction property.
\end{thm}
\vspace*{0.2cm}
\noindent  Thus Type II affords the possibility of mimicking the $\ell_0$ norm in the presence of generalized likelihoods but with the advantageous potential for drastically fewer local minima.  This is a direction for future research.  Additionally, while here we have focused our attention on classification via logistic regression, these ideas can presumably be extended to other likelihood functions provided certain conditions are met.  To the best of our knowledge, while already demonstrably successful in an empirical setting, Type II classifiers and other related Bayesian generalized likelihood models have never been analyzed in the context of sparse estimation as we have done in this section.

\section{Conclusion}

The dual-space view of sparse linear or generalized linear models naturally allows us to transition $\bfx$-space ideas originally developed for Type I and apply them to Type II, and conversely, apply $\bfgamma$-space techniques from Type II to Type I.  The resulting symmetry promotes a mutual understanding of both methodologies and helps ensure that they are not underutilized.

\newpage

\bibliographystyle{IEEEbib}

\end{document}